\documentclass[12pt,letterpaper]{article}
\usepackage{amsmath,graphicx}
\usepackage{subfigure,caption,amsfonts}
\usepackage[ruled,vlined]{algorithm2e}
\usepackage{amssymb}

\newcommand{\qed}{\rule{7pt}{7pt}}

\begin{document}
\title{Heteroscedastic Relevance Vector Machine}
\author{Daniel Khashabi, Mojtaba Ziyadi, Feng Liang\\ \{khashab2,ziyadi2,liangf\}@illinois.edu \\ University of Illinois, Urbana-Champaign}
\date{}
\maketitle
\begin{abstract}
In this work we first propose a heteroscedastic generalization to RVM, a fast Bayesian framework for regression, based on some recent similar works. We use variational approximation and expectation propagation to tackle the problem. The work is still under progress and we are examining the results and comparing with the previous works.

\end{abstract}

\section{Introduction}
The Relevance Vector Machines (RVM) \cite{tipping2001sparse} are among the most popular fast Bayesian frameworks. In this model it is assumed that the output are contaminated with a Gaussian noise with constant variance. While simple and effective in some applications this assumption is not sufficient, and we need to assume variable variance noise processes, i.e. heteroscedastic models. Gaussian processes (GP) provide a rich nonparametric tools for modelling. There are interesting similarities between GP models and RVM; in \cite{quinronero2003prediction,mackay1997gaussian} it is shown that infinite equally spaced Gaussian kernel functions will converge to a Gaussian process model. Thus one may think of RVM as a fast sparsified GP. 

Due to richness and flexibility of the GP models, there has been a surge of interests in using these model for different scenarios, especially for modelling output-heteroscedastic data.\footnote{Note that heteroscedasticity could be both in input or output, and also note that uncertainty is also different from heteroscedasticity. Examples of uncertainty uncertainty which are not heteroscedastic are \cite{tresp1994training,wright1999bayesian,dallaire2011approximate,mchutchon2011gaussian}. In addition one other incentive to create output-heteroscedastic models, are input-uncertain data, i.e. one could approximate input noise by taking its effect in the variable-variance noise in the output into account, i.e. via output-heteroscedasity.} A ground work by \cite{goldberg1997regression} assumes a model consisting of a Gaussian process for real prediction and the noise process (on the output), and evaluates the model using MCMC.  Similar models are proposed in \cite{quadrianto2009kernel} and \cite{kersting2007most}, which have slow convergence.  In \cite{lazaro2011variational} an output-heteroscedastic GP via GP noise is introduced which uses a variational approximation to train. In the same model in \cite{munoz2011heteroscedastic} the approximate training is being done using Expectation Propagation \cite{minka2001expectation}.

%Snelson and Ghahramani \cite{snelson2012variable} suggested a setting in which the importance of the pseudo-input points are variable. 
%\cite{mchutchon2011gaussian} assumes a local linear model for input noise around each training point

%As few other related works in other frameworks are, Tresp et al. \cite{tresp1994training} and Wright  \cite{wright1999bayesian} address uncertainty in input space, 

In this work follow a similar modelling to \cite{munoz2011heteroscedastic} and \cite{lazaro2011variational} to impose heteroscedasticity in RVM using a GP model noise. Due to sparsification properties of RVM model over GP, our model has faster convergence comparing to \cite{munoz2011heteroscedastic, lazaro2011variational}. The simplified variational approximation simplification which is known as \textit{collapsed variational approximation} is first introduced in \cite{king2006fast,teh2007collapsed} and analytically analyzed in \cite{hensman2012fast}.  We implement the model on several applications and compare the practical result to several existing models.  

\section{The learning model}
Consider we have the data-set $\mathcal{D} = \left\lbrace \mathbf{x}_i, y_i \right\rbrace_{i=1}^{N}  = (\mathbf{X,y})$, and $y$ is the model's scalar output, $\mathbf{x}$ is the input vector with length $Q$. Assume the following linear model similar to \cite{tipping2001sparse}, 
$$
y(\mathbf{x}) = \sum_{i=1}^{M} w_i \phi_i(\mathbf{x}) + \epsilon = \Phi(\mathbf{x})  \mathbf{w} + \epsilon 
$$
in which  $\mathbf{w} = \left[ w_1, \hdots,  w_M \right]^T $ is the vector of weights and $\Phi(\mathbf{x}) = \left[ \phi_1(\mathbf{x}), \hdots, \phi_M(\mathbf{x}) \right] $  is the vector of kernel function with length $M$.  We also define ${N \times M}$ design matrix as $\mathbf{\Phi} = \left[ \Phi(\mathbf{x}_1)^T  \hdots \Phi(\mathbf{x}_N)^T   \right]^T$ which will come handy later.  We also put normal distribution over the weights, 
$$
p(\mathbf{w} | \boldsymbol{\alpha} ) = \prod_{i=1}^{M} p( w_i | \alpha_i  ) = \prod_{i=1}^{M} \mathcal{N} \left( w_i | 0, \alpha_i^{-1} \right) = \mathcal{N} \left( \mathbf{\mathbf{w} | \mathbf{0}, \mathbf{A}^{-1}}\right). 
$$
where $\mathbf{A} = \text{diag} \left\lbrace \alpha_1, \hdots, \alpha_M \right\rbrace$.  The main difference of the above formulation with the one in \cite{tipping2001sparse} is that in the above model, the noise term has a variable variance which is modelled with a GPR similar to \cite{lazaro2011variational},
$$
 \epsilon \sim \mathcal{N}( 0, e^{g(\mathbf{x})} ) , \; g(\mathbf{x}) \sim \mathcal{GP}( \mu_0 \mathbf{I}, \mathbf{K}_g ).
$$
%The prior on $\mathbf{g}$ is as following, 
%$$
%p(\mathbf{g} | \theta_g) = \mathcal{N} \left( \mathbf{0}, \mathbf{K}_g  \right) 
%$$
$$
p \left(f_n = \Phi \left(\mathbf{x}_n \right) \mathbf{w}  |  \mathbf{A} \right) = \mathcal{N} \left( \mathbf{0} , \Phi (\mathbf{x}_n)  \mathbf{A}^{-1} \Phi^T (\mathbf{x}_n) \right) \Rightarrow  p(\mathbf{f}) = \mathcal{N} \left( \mathbf{0}, \mathbf{K}_f \right).  
$$
%In the above notation, $\mathbf{K}_f = \text{diag} \left\lbrace \Phi (\mathbf{x}_1)  \mathbf{A}^{-1} \Phi (\mathbf{x}_1) , \hdots, \Phi (\mathbf{x}_N)  \mathbf{A}^{-1} \Phi (\mathbf{x}_N)  \right\rbrace $. Note that we can write  $\Phi (\mathbf{x}_n)  \mathbf{A}^{-1} \Phi (\mathbf{x}_n)   = \sum_{i=1}^{N} \alpha_i \phi^2_i ( \mathbf{x}_n ) $.
The hyper-parameters to be learnt in this model are $\Theta = \left\lbrace \mathbf{A}, \theta_g  \right\rbrace$, in which $\theta_g$ is set of parameters in the covariance function of $g(\mathbf{x})$. 
%\begin{subequations}
%\begin{align}
%\mathcal{L}_n %& =  \frac{1}{   e^{g_n} \sqrt{2 \pi}  } \exp \left\lbrace  -\frac{ (y_n - f_n)^2  }{ 2e^{2g_n} }  \right\rbrace  \\ 
%& = \frac{1}{   e^{g_n} \sqrt{2 \pi}  } \exp \left\lbrace  -\frac{ (y_n - \mathbf{w}\Phi (\mathbf{x}_n) )^2  }{ 2e^{2g_n} }  \right\rbrace .
%\end{align}
%\end{subequations}
The likelihood is, 
\begin{subequations}
\begin{align}
\label{eq:likelihood}
\mathcal{L}   &= \prod_{n=1}^{N} \frac{1}{ e^{g_n} \sqrt{2 \pi}  } \exp \left\lbrace  -\frac{ (y_n - \Phi (\mathbf{x}_n)\mathbf{w} )^2  }{ 2e^{2g_n} }  \right\rbrace    \\
& = \frac{1}{ \exp\left[ \sum_{n=1}^{N}  g_n \right]  \left( 2 \pi \right) ^ {N/2} } \exp \left\lbrace  -\frac{1}{2} \sum_{n=1}^{N} e^{-2g_n}    \left(  y_n -   \Phi(\mathbf{x}_n) \mathbf{w}   \right)^2    \right\rbrace  \\
& = \mathcal{N} \left(  \mathbf{y} | \boldsymbol{\mu}_l , 
\mathbf{\Sigma}_l 
 \right) 
\end{align}
\end{subequations}
in which $ \mathbf{\Sigma}_l = \text{diag} \left\lbrace e^{2 g_1}, \hdots, e^{2 g_N}     \right\rbrace $ and $ \boldsymbol{\mu}_l= \left[        \Phi(\mathbf{x}_1)\mathbf{w}, \hdots, \Phi(\mathbf{x}_N)\mathbf{w}   \right]^T  = \mathbf{ \Phi w} $. 

The posterior for each $\mathbf{f}$ and $\mathbf{g}$ is, 
\begin{equation}
p(\mathbf{g} |  \mathbf{w}, \mathcal{D}, \Theta ) \propto \mathcal{N} \left( \mu_0 \mathbf{I}, \mathbf{K}_g \right) \times \mathcal{L}
\label{eq:nonGaussian:post}
\end{equation}
\begin{equation}
p(\mathbf{w} |  \mathbf{g}, \mathcal{D}, \Theta ) \propto \mathcal{N} \left( \mathbf{0}, \mathbf{A}^{-1} \right) \times \mathcal{L}
\end{equation}
Note that the above likelihood is only Gaussian with respect to $\mathbf{w}$, not $\mathbf{g}$. Considering that both priors are Gaussian, this causes the posterior of $\mathbf{g}$ in Eq.\ref{eq:nonGaussian:post} to become non-Gaussian, and need to approximated to become tractable. \\
Also note that defining $\mathbf{f} = \Phi \mathbf{w}$, because of the deterministic relationship between $\mathbf{f}$ and $\mathbf{w}$, having one, entails having the others distribution, and the model resulting model based on $\mathbf{f}$ is essentially the same as the one in \cite{lazaro2011variational}. To keep everything simple here instead of writing the Bayes relation on $\mathbf{f}$, we simply write it on $\mathbf{w}$. 

\section{Approximate learning}
In RVM training the parameters is achieved via maximizing the marginal likelihood, $p(\mathbf{y} | \mathbf{X},  \mathbf{A}, \sigma^2 )$, in which $\sigma^2$ is a constant variance. While in out setting the variance $e^{g(\mathbf{x})}$ is not constant, and it not determined.

In this section we formulate the approximations to tackle the problem at hand. We'll first introduce the variational methods and then expectation propagation method. 
\subsection{Variational Approximation}
In variable approximation we approximate the posterior distribution by factorizing the joint distribution of target parameters into independent distributions. We then lower bound the marginal likelihood, henceforth instead of maximizing the posterior, we maximize the lower bound. A comprehensive review of variational methods for several applications could be found in \cite{wainwright2008graphical}. We assume our approximate factorization as follows, 
$$
p(\mathbf{w,g} | \mathbf{y}) \approx q(\mathbf{w,g})  =  q(\mathbf{w})  . q(\mathbf{g})     
$$
The optimization of the above independence approximation could be done using Kullback-Leibler divergence, $\text{KL}\left( q\left(\mathbf{w}\right)  . q\left(\mathbf{g} \right) \| p\left(\mathbf{w,g} | \mathbf{y} \right)    \right) $, which at the same time maximizes the lower bound on the marginal likelihood, 
$$
\log p(\mathbf{y}) \geq \int q(\mathbf{g}) q(\mathbf{w}) \log \frac{p(\mathbf{w,g}, \mathbf{y})}{q(\mathbf{g}) q(\mathbf{w})} d \mathbf{w} d \mathbf{g}  = \mathcal{F}\left(q(\mathbf{w}), q(\mathbf{g})  \right) 
$$
which can be rewritten as 
$$
\log p(\mathbf{y}) -  \mathcal{F}\left(q(\mathbf{w}), q(\mathbf{g})  \right) =  \text{KL}\left( q\left(\mathbf{w}\right)  . q\left(\mathbf{g} \right) \| p\left(\mathbf{w,g} | \mathbf{y} \right)    \right) 
$$
The solution to the about variational minimization is 
$$
q^{t+1}(\mathbf{w}) \propto  p(\mathbf{w}) . \exp \left[  \int q^{t}(\mathbf{g}) \log  p( \mathbf{y} | \mathbf{w,g}) d \mathbf{g}  \right]  
$$
$$
q^{t+1}(\mathbf{g}) \propto  p(\mathbf{g}) . \exp \left[  \int q^{t+1}(\mathbf{w}) \log  p( \mathbf{y} | \mathbf{w,g}) d \mathbf{w}  \right]  
$$
As suggested in \cite{lazaro2011variational} the above lower bound $\mathcal{F}$, could be more simplified by exploiting the structure of the problem.  It turns out that this method was previously introduced by \cite{king2006fast,teh2007collapsed} and known as \textit{collapsed variational} inference. The recent work \cite{hensman2012fast} introduces a unified view of this model. Similarly  we can analytically marginalize  one variable and put in the lower bound to decrease a portion of the optimization process. We analytically compute $q(\mathbf{w})$,
\begin{equation}
\label{eq:q:star}
q^*(\mathbf{w}) = \arg \max_{q(\mathbf{w})} = \frac{p(\mathbf{w})}{ \mathcal{Z} \left(  q \left( \mathbf{w} \right)  \right) } \exp  \left[ \int    q(\mathbf{g})  \log p(  \mathbf{y} |  \mathbf{w,g} ) d \mathbf{g} \right],
\end{equation}
where $  \mathcal{Z} \left(  q \left( \mathbf{w} \right)  \right) = \int  p(\mathbf{w})  \exp \left[  \int  q(\mathbf{g})  \log p(  \mathbf{y} |  \mathbf{w,g} ) d \mathbf{g} \right]  d \mathbf{w}$ is the normalizing constant. Now we simplify the lower bound: 
\begin{subequations}
\begin{align}
\log p( \mathbf{y} )  \geq \left. \mathcal{F} \right|_{q(\mathbf{w}) = q^*(\mathbf{w}) } & = \left. \mathbb{E}_{ \mathbf{w,g} \sim q }  \log \frac{ p(\mathbf{w,g,y})  }{  q(\mathbf{w})  q(\mathbf{g})  } \right|_{q(\mathbf{w}) = q^*(\mathbf{w}) }   \\
& =  \mathbb{E}_{ \mathbf{w} \sim  q^*  }  \mathbb{E}_{ \mathbf{g} \sim q }   \log \frac{ p(\mathbf{w,g,y})  \mathcal{Z}  }{  q(\mathbf{g})  p(\mathbf{w}) \exp \left[ \mathbb{E}_{ \mathbf{g} \sim q }  \log p(\mathbf{y |  w, g} )  \right]  }   \\
& = \mathbb{E}_{ \mathbf{w} \sim  q^*  }  \mathbb{E}_{ \mathbf{g} \sim q } \left[   \log \frac{ p(\mathbf{y | w, g})  p(\mathbf{w}) p(\mathbf{g})   \mathcal{Z}  }{  q(\mathbf{g})  p(\mathbf{w})   }  -  \mathbb{E}_{ \mathbf{g} \sim q }  \log p(\mathbf{y |  w, g} )   \right]  \\
& = \mathbb{E}_{ \mathbf{w} \sim  q^*  }  \mathbb{E}_{ \mathbf{g} \sim q } \left[   \log \frac{  p(\mathbf{g})   \mathcal{Z}  }{  q(\mathbf{g})    }     \right]  \\
& = \mathbb{E}_{ \mathbf{g} \sim q } \left[   \log \frac{  p(\mathbf{g})   \mathcal{Z}  }{  q(\mathbf{g})    }     \right]  \\
& =\mathcal{Z}  -  \mathbb{E}_{ \mathbf{g} \sim q } \left[   \log \frac{  p(\mathbf{g})  }{  q(\mathbf{g})    }     \right]   = \mathcal{Z}  - \text{KL}\left(   q(\mathbf{g}) \|  p\left(\mathbf{g}\right)  \right)  = \mathcal{F}_{\text{KL}}
\end{align}
\end{subequations}
Which is what we expected, a new variational bound which is functional of only the distribution on $\mathbf{g}$. Similar to \cite{hensman2012fast} we call this new bound $\mathcal{F}_{\text{KL}}$. Based on the assumptions of the problem now we can simplify the bound. Following \cite{lazaro2011variational} we assume that $q(\mathbf{g}) = \mathcal{N}  \left( \mathbf{g}  | \boldsymbol{\mu}, \mathbf{\Sigma}  \right) $, 

\begin{subequations}
\begin{align}
\mathcal{F}_{\text{KL}} & =  \log \int \mathcal{N} \left( \mathbf{w} | \mathbf{0}, \mathbf{A}^{-1}  \right)  \exp \left[ \int \mathcal{N} (  \mathbf{g} |  \boldsymbol{\mu}, \mathbf{\Sigma}  )  \log p( \mathbf{y | w,g} )  d \mathbf{g}  \right]   d\mathbf{w} \\  
& \qquad  - \text{KL} \left(  \mathcal{N}(   \mathbf{g} |  \boldsymbol{\mu}, \mathbf{\Sigma}  )  \|    \mathcal{N}  ( \mathbf{g} |  \mu_0 \mathbf{1}, \mathbf{K}_g ) \right)
\end{align}
\end{subequations}
Substituting the term inside the exponential $ \int \mathcal{N}(  \mathbf{g} |  \boldsymbol{ \mu }, \boldsymbol{\Sigma}  )  \log p(\mathbf{y | w,g} ) d \mathbf{g}  = \log \mathcal{N}(\mathbf{y} |  \mathbf{w}, \mathbf{R} ) - \frac{1}{4} \text{tr} (\mathbf{\Sigma})$ in which $[\mathbf{R}]_{ii} = e^{  [\boldsymbol{\mu}]_i - [ \mathbf{\Sigma} ]_{ii} /2   }$, which makes the bound as follows: 
\begin{equation}
\mathcal{F}_{\text{KL}} = \log \mathcal{N}(  \mathbf{y} | \mathbf{0}, \mathbf{\Phi} \mathbf{A}^{-1}\mathbf{\Phi}^T  +  \mathbf{R}  ) - \frac{1}{4} \text{tr}(\mathbf{\Sigma}) - \text{KL}( \mathcal{N}(\mathbf{g} |  \boldsymbol{\mu}, \mathbf{\Sigma}  )  \| \mathcal{N} ( \mathbf{g} | \mu_0 \mathbf{1} , \mathbf{K}_g  )  ).
\label{eq:lowerboundNEw}
\end{equation}
Similar to what is done in \cite{lazaro2011variational} one could exploit the nature of the formulation and decrease the number of the problems to be optimized by using equations of $\frac{\partial \mathcal{F}_{\text{KL}} }{\partial \boldsymbol{\mu}  } = 0$, and $\frac{\partial \mathcal{F}_{\text{KL}} }{\partial \boldsymbol{\Sigma}  } = 0$, and expressing the  unknown variables in terms of one single smaller variable $\Lambda$.  Simplifying the update equations will result in the same equation as in appendix of \cite{lazaro2011variational}, but with more straightforward equation for updating $\boldsymbol{\alpha}$ parameters,
$$
\frac{\partial \mathcal{F}_{\text{KL}}  }{ \partial \alpha_j  } = 0 \Rightarrow \alpha_j = \sum_{k=1}^{N} \frac{  \phi_j ( \mathbf{x}_k ) }{  {y_k}  - \exp\left\lbrace  [\boldsymbol{\mu}]_{k} - [\boldsymbol{\Sigma}]_{kk}/2 \right\rbrace  - \sum_{i=1, i \neq j}^{N} \frac{1  }{ \alpha_i } \phi_i(\mathbf{x}_k)  }
$$

%\begin{observation}
%One could easily incorporate variable scale kernels. 
%\end{observation}

\subsection{Expectation Propagation Approximation}
One could use \cite{minka2001expectation} to do iterative approximation to $\mathbf{g}$'s posterior. We need to approximate the likelihood in Eq. \ref{eq:likelihood} into factors using a Gaussian family, 
$$
p(y_n |  \mathbf{w}, \mathbf{g} ) \approx \tilde{t}_{y_n} = \tilde{Z}. \mathcal{N}(y_n |  \tilde{\mu}_{y_n},  \tilde{\sigma}^2_{y_n}   ), 
$$
The likelihood is, 
$$
\mathcal{L} = p(\mathbf{y} |  \mathbf{w}, \mathbf{g}) \approx \prod_{n=1}^{N} \tilde{t}_{y_n} =  \mathcal{N}(\mathbf{y} |  \tilde{\boldsymbol{\mu}}_{y},  \tilde{\mathbf{\Sigma}}_{y} ). \prod_{n=1}^{N}\tilde{Z} = \mathcal{L}_{EP}
$$
Putting the prior and the likelihood into the Bayes formula in Eq. \ref{eq:nonGaussian:post} we get the following relation for posterior, 
$$
q(\mathbf{g} | \mathbf{w}, \mathcal{D}, \Theta) \propto p(\mathbf{q}) \times \mathcal{L}_{EP} 
$$
Such that $q(\mathbf{g} | \mathbf{w}, \mathcal{D}, \Theta) = \mathcal{N}(\mathbf{g} | \boldsymbol{\mu}, \mathbf{\Sigma} ) $ in which $\mathbf{\Sigma} = \left( \mathbf{K}^{-1}_g+ \tilde{\mathbf{\Sigma}}^{-1}_{y} \right)^{-1}$ and $\boldsymbol{\mu} = \mathbf{\Sigma} \tilde{\mathbf{\Sigma}}^{-1} \tilde{\mathbf{\mu}} $. The normalizing constant for the above Bayes relation, $\mathcal{Z}_{EP}$ is the EP approximation for marginal likelihood.\\
In EP at each step we delete the $i$-th term from the posterior to get the \textit{gap distribution}, and then we add the corresponding exact likelihood to the gap distribution. Then we minimize KL(,) for the resulted non-Gaussian distribution and a Gaussian distribution, and then based on the approximated Gaussian, we update $\tilde{t}_{y_n}$.

\section{Prediction}
Note that using Eq. \ref{eq:q:star}  we have 
$$
q^*(\mathbf{w}) \propto \mathcal{N}(\mathbf{0}, \mathbf{A}^{-1}). \mathcal{N}(\mathbf{y} | \mathbf{f,R} )  = \mathcal{N}\left( \boldsymbol{\mu}_w, \boldsymbol{\Sigma}_w \right) 
$$
%  \mathbf{A}^{-1}, \mathbf{A}^{-1} - \left( \mathbf{A}^{-1} + \mathbf{R} \right)   \mathbf{A}^{-1} \right) 
$$
 \boldsymbol{\mu}_w = \mathbf{\Sigma}_w \mathbf{ \Phi}^T  \mathbf{ R}^{-1} \mathbf{y}, \;   \boldsymbol{\Sigma}_w = \left( \mathbf{A + \Phi }^{T} \mathbf{R}^{-1} \mathbf{\Phi }    \right) ^{-1} 
$$
Comparing the above posterior mean and variance with that of Eq.6 in \cite{tipping2003fast} it reveals that the matrix $\mathbf{R}$ here acts like the $\sigma^2\mathbf{I}$ noise matrix in RVM.
Now we can find the predictive distribution by marginalizing the latent variables in model. 
%
%Similar to classic RVM we can find the predictive distribution could be found 
%$$
%p(y_* | \mathbf{x}_*, \mathcal{D}, \Theta ) = \int_{\mathbf{g}} \int_{\mathbf{w}} p( y_* | \mathbf{x}_*, \mathcal{D}, \Theta, \mathbf{w,g}  ) p(\mathbf{})  
%$$
%Note that unlike \cite{} the matrices inside inverse are all diagonal and easy to compute. 
%$$
%q(w^*) = \int p(w^* | \mathbf{x}^*, \mathcal{D}, \Theta  ). q(\mathbf{w}) d\mathbf{w}  = \mathcal{N} \left(w^* |  \mu_{w^*}, \Sigma_{w^*} \right) 
%$$
%The same could be done to calculate $q(g^*) $ and $q(y^*) $.  
%
%We can find the predictive values for the noise using standard Gaussian process formulation: 
%$$
%q(g_*)  = \int p(g_* | \mathbf{x}_*, \mathcal{D}, \mathbf{g}) q(\mathbf{q}) d \mathbf{g}  =  \mathcal{N} \left(g_* | \mu_{g_*}, \sigma^2_{g_*}  \right)  
%$$
%$$
%\mu_{g_*} = , \; \sigma^2_{g_*} = 
%$$
\section{Notes on convergence}
In conventional RVM, the prior $p(  \mathbf{w}  | \boldsymbol{\alpha})$ acts to sparsify the model. During the update procedure of the model, some $\alpha_i \rightarrow \infty $, then we put the corresponding $w_i = 0$, and continue the training, and results in sparsity of RVM. In \cite{tipping2003fast} it's been showed that the local maxima of marginal likelihood for some values of $\alpha_i$ happens in infinity.  

Assume the lower bound in Eq.\ref{eq:lowerboundNEw}; in this bound the first part is a function of $\boldsymbol{\alpha}$ and need to be optimized. Comparing this with the likelihood in \cite{tipping2003fast}( Eq.7 ) we see that the factor $\mathbf{C} = \sigma^2 \mathbf{I} + \mathbf{\Phi A} ^{-1} \mathbf{\Phi}^{T}$ defined in that work is equivalent to $\mathbf{C}^{\prime} = \mathbf{R}+ \mathbf{\Phi A} ^{-1} \mathbf{\Phi}^{T}$.  Thus, similarly we could follow the Eq.15 to Eq.21  in  \cite{tipping2003fast} to show that the local maxima of our lower bound happens when some $\alpha_i$ go to infinity. So in experiments for a big enough $\alpha_{Th}$, for each $\alpha_i > \alpha_{Th}$ we could put $w_i = 0$  and get smaller model. 

Also note that due to sparsification properties of RVM, our model has faster convergence comparing to \cite{munoz2011heteroscedastic, lazaro2011variational} as we calculate $\boldsymbol{\alpha}$ parameters analytically.

%\section{Healing RVM}
%In \cite{rasmussen2005healing}  it is pointed out that the predictive uncertainty in RVM, unintuitively, gets smaller as we get away from the training samples, i.e. decreases to noise level.  The aforementioned work has proposed to ?? to heal this problem of RVM. 
%
%\section{Experiments}

\section{Conclusion and future work}
We derived formulations of Heteroscedastic RVM which are similar to \cite{lazaro2011variational} and has interesting similarities with \cite{tipping2003fast}. This is a report of an ongoing work, and we are still experimenting and comparing methods. The final results will soon be published. 
%One obvious future work is the approximation used; we try to investigate for fast while more accurate approximations. Another possible future work is\\ 

\bibliographystyle{apalike}

\bibliography{ref}

\end{document}